\documentclass[journal]{IEEEtran}
\usepackage{amsmath,graphicx}
\usepackage{color,hyperref}
\usepackage[table,xcdraw]{xcolor}

\usepackage{times}
\usepackage{epsfig}
\usepackage{graphicx}
\usepackage{amsmath}
\usepackage{amssymb}

\usepackage{algorithm}
\usepackage{algorithmicx}
\usepackage{algpseudocode}
\usepackage{multirow}
\usepackage{booktabs} 
\usepackage{url}
\usepackage{cite}
\usepackage{bm}
\usepackage{booktabs}
\usepackage{makecell}
\usepackage{xcolor}

\newcommand{\R}{\mathbb{R}}

\usepackage{hyperref}
\hypersetup{
    colorlinks=true,
    linkcolor=blue,
    filecolor=blue,
    urlcolor=blue
    }

\definecolor{blue1}{rgb}{0, 0.439, 0.753}
\definecolor{red}{rgb}{0.753, 0, 0}

\newlength\savedwidth

\begin{document}
%
\title{Change Detection Meets Visual Question Answering}
\author{Zhenghang Yuan,~\IEEEmembership{Student Member,~IEEE,} Lichao Mou,
	Zhitong Xiong, ~\IEEEmembership{Member,~IEEE,}
	and \\Xiao Xiang Zhu,~\IEEEmembership{Fellow,~IEEE}

\IEEEcompsocitemizethanks{This work is jointly supported by the China Scholarship Council, by the European Research Council (ERC) under the European Union's Horizon 2020 research and innovation programme (grant agreement No. [ERC-2016-StG-714087], Acronym: \textit{So2Sat}), by the Helmholtz Association through Helmholtz Excellent Professorship ``Data Science in Earth Observation - Big Data Fusion for Urban Research'' (grant number: W2-W3-100), by the German Federal Ministry of Education and Research (BMBF) in the framework of the international future AI lab ``AI4EO -- Artificial Intelligence for Earth Observation: Reasoning, Uncertainties, Ethics and Beyond'' (grant number: 01DD20001) and by the German Federal Ministry for Economic Affairs and Climate Action in the framework of the ``national center of excellence ML4Earth'' (grant number: 50EE2201C). 


Z. Yuan, L. Mou, Z. Xiong and X. X. Zhu are with the Data Science in Earth Observation, Technical University of Munich (TUM), 80333 Munich, Germany. 
(e-mails: zhenghang.yuan@tum.de; lichao.mou@tum.de; zhitong.xiong@tum.de; xiaoxiang.zhu@tum.de)

}
}

\markboth{}%
{Shell \MakeLowercase{\textit{et al.}}: Bare Demo of IEEEtran.cls for IEEE Journals}

\maketitle

\begin{abstract}
The Earth's surface is continually changing, and identifying changes plays an important role in urban planning and sustainability. Although change detection techniques have been successfully developed for many years, these techniques are still limited to experts and facilitators in related fields. In order to provide every user with flexible access to change information and help them better understand land-cover changes, we introduce a novel task: change detection-based visual question answering (CDVQA) on multi-temporal aerial images.
In particular, multi-temporal images can be queried to obtain high level change-based information according to content changes between two input images. We first build a CDVQA dataset including multi-temporal image-question-answer triplets using an automatic question-answer generation method. Then, a baseline CDVQA framework is devised in this work, and it contains four parts: multi-temporal feature encoding, multi-temporal fusion, multi-modal fusion, and answer prediction. In addition, we also introduce a change enhancing module to multi-temporal feature encoding, aiming at incorporating more change-related information. Finally, effects of different backbones and multi-temporal fusion strategies are studied on the performance of CDVQA task. The experimental results provide useful insights for developing better CDVQA models, which are important for future research on this task. The dataset will be available at {https://github.com/YZHJessica/CDVQA}.
\end{abstract}

\section{Introduction}
\label{sec:intro}
\IEEEPARstart{T}HE Earth's surface is continually changing by man-made and natural influences. These changes are closely involved in human and social development and also guide the urban planning and sustainability \cite{saha2020building}. Change detection, aiming at detecting differences of the same region at different times, has become a research priority in recent decades \cite{ban2016change,you2020survey,leenstra2021self}. Timely and effective change information can be used for many practical applications such as environmental management \cite{tian2014improving,baker2007change,schmitt2013wetland}, natural disasters monitoring \cite{washaya2018coherence,qiao2020novel}, urban land-use \cite{olteanu2020use,mishra2014sensitivity} and agriculture production \cite{haack1998remote}. 

Nowadays, change detection technology has been developed significantly, and there are various algorithms with great performance improvement for remote sensing data \cite{liu2019review,du2019unsupervised,li2019unsupervised}. Change detection methods can be divided into two main categories, depending on whether or not the types of changes are detected. 
One is binary change detection that only detects changed regions but ignores the type of changes, e.g., the object-oriented key point vector distance for detecting binary land-cover changes \cite{lv2020object} and the end-to-end 2D CNN for hyperspectral image change detection \cite{wang2018getnet}. Change maps obtained by such methods are visualized by binary values to depict change information at the pixel level. The other is semantic change detection, for instance, using asymmetric Siamese network for identifying changes via feature pairs \cite{yang2020asymmetric} and reasoning bi-temporal semantic correlations \cite{ding2021bi}. These methods not only detects changed regions but also identify change types.
 

\begin{figure}
	\centering
	\includegraphics[width=0.5\textwidth]{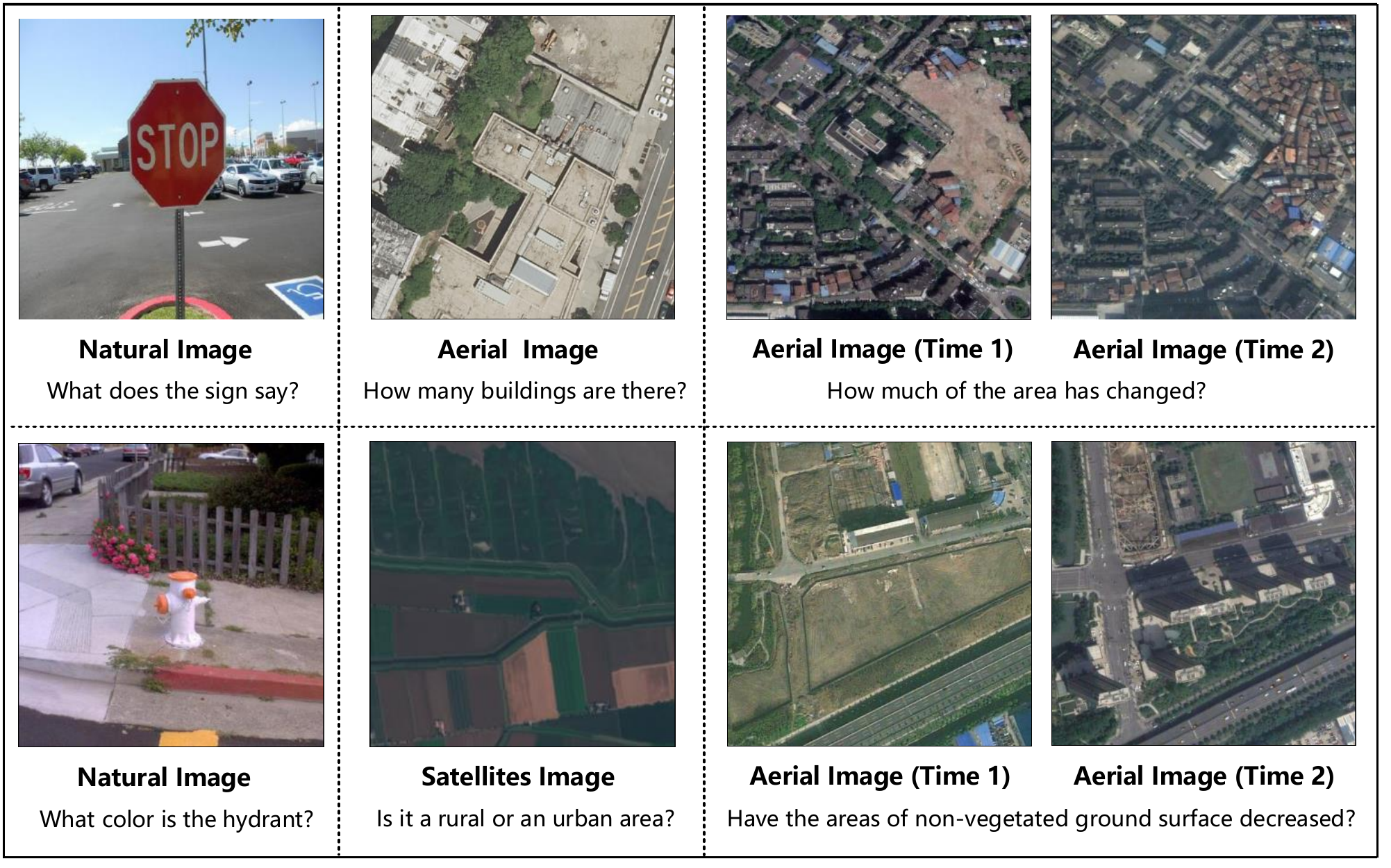}
	\caption{Examples of questions for natural imagery, aerial imagery, and multi-temporal aerial images in VQA tasks. }
	\label{motivation}
\end{figure}

Although change detection has great application value, the specialized nature of this task makes change information limited to researchers. It is still difficult for end users to access and understand much of important change information. For instance, ordinary users are interested in a certain change type in a certain region, but it is inconvenient and ineffective for them to find it on change maps in practical applications. Considering this problem, efficient and effective change information interaction with end users becomes important. 
In this context, nature language processing (NLP) enables computers to understand the text in almost the same way as humans. It is user-friendly and can greatly improve the interactivity between image analysis systems and end users. Therefore, in order to alleviate the interaction issue, the integration of computer vision and NLP \cite{mogadala2019trends} has gradually become a hot research topic in the machine learning community. In particular, tasks like visual description generation \cite{khamparia2020integrated}, visual storytelling \cite{ferraro2016visual}, visual question answering (VQA) \cite{antol2015vqa,wu2017visual} and visual dialog \cite{tu2021learning} have been fully and successfully conducted in computer vision. Similarly, tasks of integrating remote sensing imagery and NLP, such as image captioning and VQA, have also become an active research topic in the field of remote sensing \cite{zheng2021mutual, yuan2021self}. Captioning for remote sensing images was first proposed in \cite{qu2016deep}, and Lu et al. \cite{lu2017exploring} further explored captioning methods using both handcrafted and convolutional features and proposed a new dataset. Recently, a multilayer aggregated Transformer was utilized to extract information for caption generation \cite{liu2022remote}. Regarding VQA for remote sensing data (RSVQA), Lobry et al. \cite{lobry2020rs} first introduced this task, built two datasets, and used a hybrid CNN-RNN model to extract feature, and Yuan et al. \cite{yuan2022easy} proposed a self-paced curriculum learning based model trained from easy to hard questions gradually.


Compared to natural images, aerial images are more specialized due to the top-view perspective and complicated background. As shown in Fig. \ref{motivation}, answers to questions about natural images \cite{antol2015vqa} are more obvious than answers to questions about aerial images \cite{lobry2020rs} in many cases in VQA tasks. Besides, Fig. \ref{motivation} illustrates that answers to questions about the comparison of multi-temporal aerial images require careful observation and even calculation, which is unfriendly to ordinary users. 
Though VQA for natural images has been studied for many years and VQA for remote sensing data has also gradually become a research focus, VQA for change detection based on multi-temporal images is under-explored.
Considering the significance of change detection task and its values in practical applications, it is vital to investigate how to improve the friendliness and accessibility of change detection systems to end users. Hence, there is also a greater need to develop end user accessible VQA algorithms for multi-temporal remotely sensed data.

In this paper, we introduce the task of change detection-based visual question answering (CDVQA) on multi-temporal aerial images. Specifically, given two aerial images captured at different times and a natural language question about them, the CDVQA task aims to provide an answer in natural language by comparing the content of two images. To this end, we create a CDVQA dataset by an automatic generation method, which contains 2,968 pairs of multi-temporal images and more than 122,000 question-answer pairs. The questions are carefully designed to cover various types of changes. Moreover, we propose a baseline method for CDVQA task as shown in Fig. \ref{mainarch}. To sum up, the main contributions of this work are summarized as follows:
\begin{figure*}
	\centering
	\includegraphics[width=0.95\textwidth]{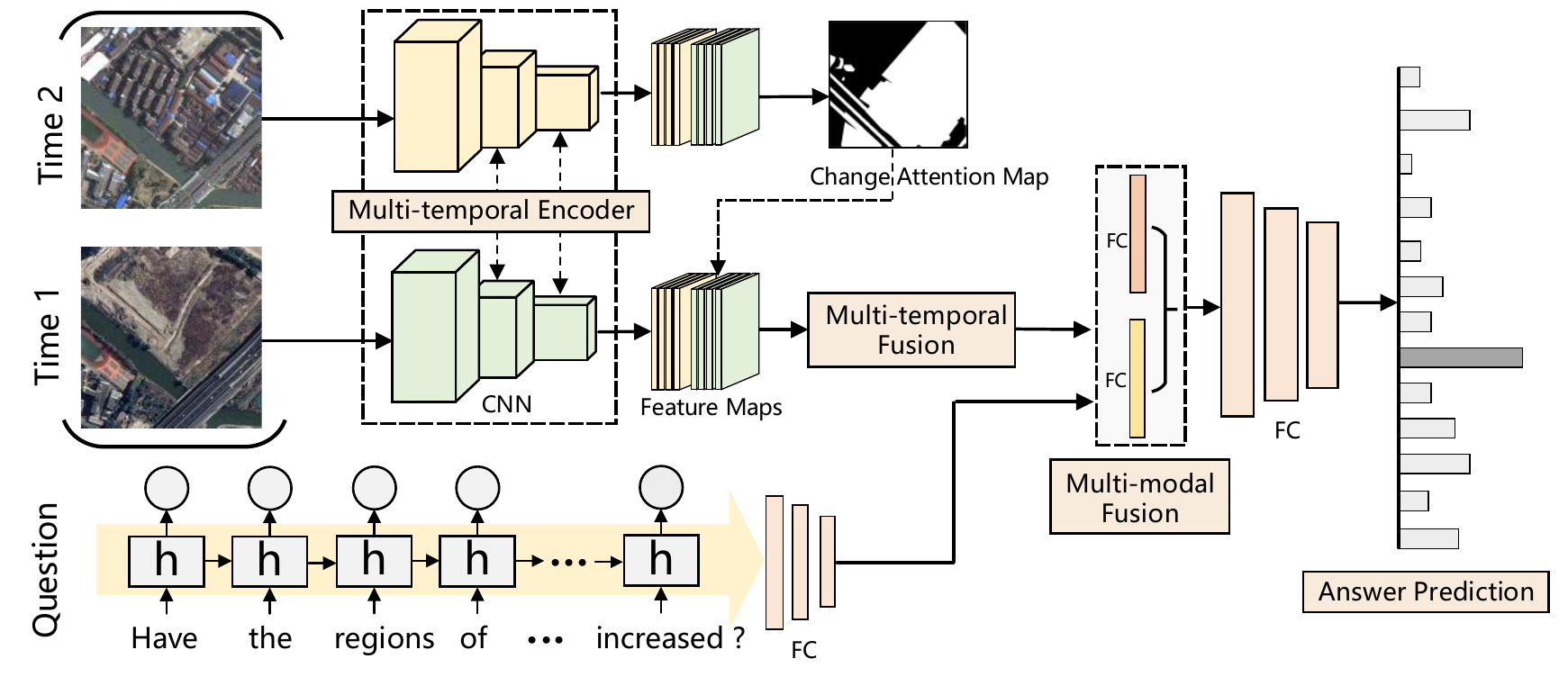}
	\caption{The main architecture of the proposed CDVQA framework. It contains four main parts: multi-temporal feature encoding, multi-temporal fusion, multi-modal fusion, and answer prediction.}
	\label{mainarch}
\end{figure*}

\begin{itemize}
	\item We design an automatic question-answer generation method and create a new CDVQA dataset. Specifically, the proposed dataset contains 2,968 pairs of aerial images and more than 122,000 corresponding question-answer pairs.
	\item A baseline framework for CDVQA task is proposed, and it includes four parts: multi-temporal feature encoding, multi-temporal fusion, multi-modal fusion and answer prediction. In addition, a change enhancing module is proposed to incorporate more change-related information into visual features. 
	\item Extensive experiments have been conducted to study effects of different network parts on the CDVQA performance. Particularly, different backbones and multi-temporal fusion strategies are investigated. The results provide useful insights on the CDVQA task.
\end{itemize}

The rest of the paper is organized as follows. The detailed information for the construction of CDVQA dataset is introduced in Section \ref{Dataset}. Section \ref{Methodology} presents the methodology. Experimental results and discussion are shown in Section \ref{Experiments}. Finally, this paper is concluded in Section \ref{Conclusion}.

\section{Dataset}
\label{Dataset}
Different from the traditional VQA task, CDVQA involves multi-temporal aerial images and requires time series analysis. Taking this into account, we choose the existing semantic change detection dataset SECOND \cite{yang2020asymmetric} as the basic data to automatically generate a CDVQA dataset. The SECOND dataset collects bi-temporal high resolution optical (RGB) images by several different aerial platforms and sensors, with spatial resolution varying from 0.5 m to 3 m \cite{ding2021bi}.  Geographical positions include several cities in China, such as Shanghai, Hangzhou, and Chengdu. It has 4,662 pairs of aerial images with the size of $512 \times 512$ pixels, and 2,968 pairs are publicly available. Each pair consists of a pre-event aerial image and a post-event image of the same location at different times. Besides, each pair has two labeled semantic change maps at the pixel level, before and after the change. In each semantic change map, non-change region and 6 land-cover classes related to changes including non-vegetated ground (NVG) surface, buildings, playgrounds, water, low vegetation and trees are annotated. The authors of the SECOND dataset declare in their paper that semantic change maps in this dataset are labeled by a team of experts in Earth vision applications and high accuracy is guaranteed. Therefore, the generated question-answer pairs in this work are highly relevant to the content of image pairs.
Overall, this dataset has critical semantic change information of main land-cover classes at pixel-level, which provides sufficient information for generating question-answer pairs for the CDVQA task. In this case, we use the 2,968 openly available pairs as our basic data for the further dataset construction.

\subsection{Multi-temporal Image-Question-Answer Triplets Construction}

Formally, in each pair of multi-temporal aerial images, let $\bm{x}_{t_1} \in \R^{3\times H\times W}$ be the image at time $\text{T}_1$, and $\bm{x}_{t_2} \in \R^{3\times H\times W}$ be the image at time $\text{T}_2$. $\bm{s}_{t_1} \in \R^{H\times W}$ and $\bm{s}_{t_2} \in \R^{H\times W}$ denote semantic change maps of $\bm{x}_{t_1}$ and $\bm{x}_{t_2}$, respectively, and each pixel in $\bm{s}_{t_1}$ and $\bm{s}_{t_2}$ indicates one semantic class, ranging from 0 to 6. Semantic change maps show changed regions and provide their change types at the pixel level. Background pixels mean non-change regions, which are the same in both ${\bm{s}_{t_1}}$ and ${\bm{s}_{t_2}}$ for an image pair. Foreground pixels indicate changed regions of different land-cover types. Specifically, the value of the pixel in ${\bm{s}_{t_1}}$ indicates the semantic class at $\text{T}_1$ and the value of the pixel in  ${\bm{s}_{t_2}}$ indicates the semantic class at $\text{T}_2$. The main advantage of introducing semantic change maps is that we can access more details about changes, i.e., we are able to know not only where changes happen but also what types they are. In this work, our motivation is to use natural language as queries to obtain these two types of information.
Given semantic change information of $\bm{s}_{t_1}$, $\bm{s}_{t_2}$, the following five types of questions are designed in the proposed dataset: \emph{change or not}, \emph{increase/decrease or not}, \emph{change to what}, \emph{largest/smallest change}, and \emph{change ratio}. In our case, the smallest/largest change refers to the land cover class that has the least/most pixels changing. These questions are of great interest to end users in real-world applications. In what follows, a detailed description of the automatic generation of multi-temporal image-question-answer triplets for different question types is given.



\begin{itemize}
\item \textbf{Change or not}. 

\emph{Change or not for an image pair}. The most fundamental yet important information in change detection is about whether a certain land-cover has changed. Note that a change occurs regardless of whether the area of a land-cover increases or decreases. For each pair of aerial images, the set of changed land-cover classes $\mathcal{L}_{t_1}$ and $\mathcal{L}_{t_2}$ are extracted from $\bm{s}_{t_1}$ and $\bm{s}_{t_2}$, respectively.
Let ${l_i}$ be a land-cover class, ${{l_i} \in {\mathcal{L}_{t_1}}}$ or ${{l_i} \in {\mathcal{L}_{t_2}}}$, indicating that the corresponding land-cover type has changed.  
In this case, the answer should be \emph{yes}. On the contrary, if ${l_i} \notin {\mathcal{L}_{t_1}}$ and ${l_i} \notin {\mathcal{L}_{t_2}}$, it indicates that the corresponding land-cover does not change. Then, the answer should be \emph{no}. All land-cover types are traversed to generate multiple question-answer pairs. 

\emph{Change or not for a single image}. For change detection tasks, sometimes one want to focus not only on whether a certain land-cover class has changed, but also on whether changes have occurred in the pre-event image or post-event image. Therefore, we extract semantic change information solely from the first or second image to generate relevant questions and answers. Please note that in this work, the first image in the image pair refers to the pre-event/pre-change, and the second image means the post-event/post-change. Particularly, for the land-cover class ${l_i}$, if ${{l_i} \in {\mathcal{L}_{t_1}}}$, it indicates that the corresponding land-cover has changed on the pre-event image. The answer under this situation should be \emph{yes}.
Similarly, if ${{l_i} \in {\mathcal{L}_{t_2}}}$, it means that the area of ${l_i}$ has changed on the post-event image. The answer will also be \emph{yes}. In other cases, i.e. ${l_i} \notin {\mathcal{L}_{t_1}}$ and ${l_i} \notin {\mathcal{L}_{t_2}}$, the corresponding answer to the question about whether it has changed on a single image should be \emph{no}.

\item \textbf{Increase/decrease or not}.

Change detection in real-world applications often requires more specific change information, for instance, whether the area of a land-cover has increased or decreased. In this context, we denote the area of ${l_i}$ in ${\bm{s}_{t_1}}$ as ${\bm{A}_{t_1}^{i}}$ and the area in ${\bm{s}_{t_2}}$ as ${\bm{A}_{t_2}^{i}}$. 
For increasing-related question-answer pairs, if ${\bm{A}_{t_2}^{i}}-{\bm{A}_{t_1}^{i}>0}$, the area of ${l_i}$ increases. Then the answer to this question should be \emph{yes}. For decreasing-related pairs, the generation process is similar. If ${\bm{A}_{t_2}^{i}}-{\bm{A}_{t_1}^{i}<0}$, the area of ${l_i}$ decreases.
Note that the area of ${l_i}$ is defined as all pixels with label $l_i$ in the whole imagery.
\begin{figure*}
	\centering
	\includegraphics[width=1.0\textwidth]{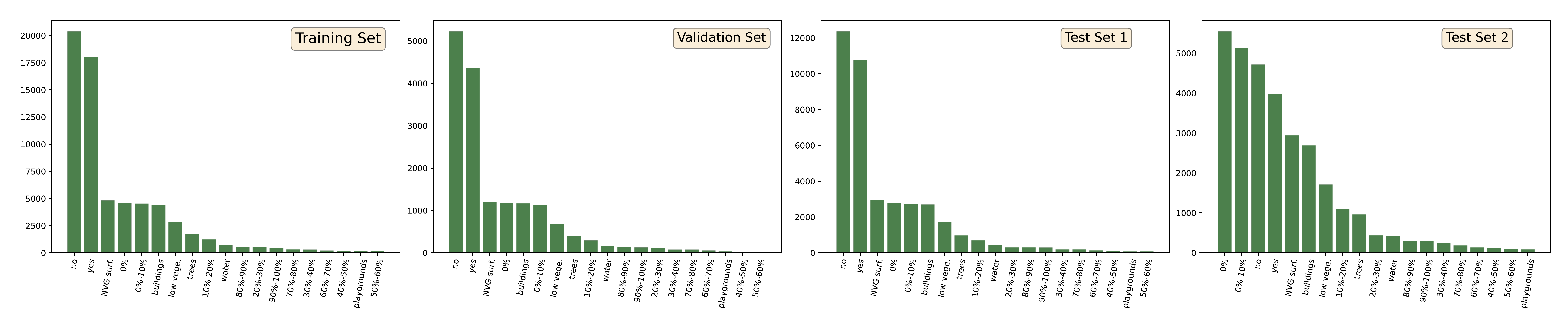}
	\caption{Visualization of answer distributions of different subsets. From left to right: training set, validation set, test set 1, and test set 2.}
	\label{answer_type_distribution}
\end{figure*}

\begin{figure}
	\centering
	\includegraphics[width=0.5\textwidth]{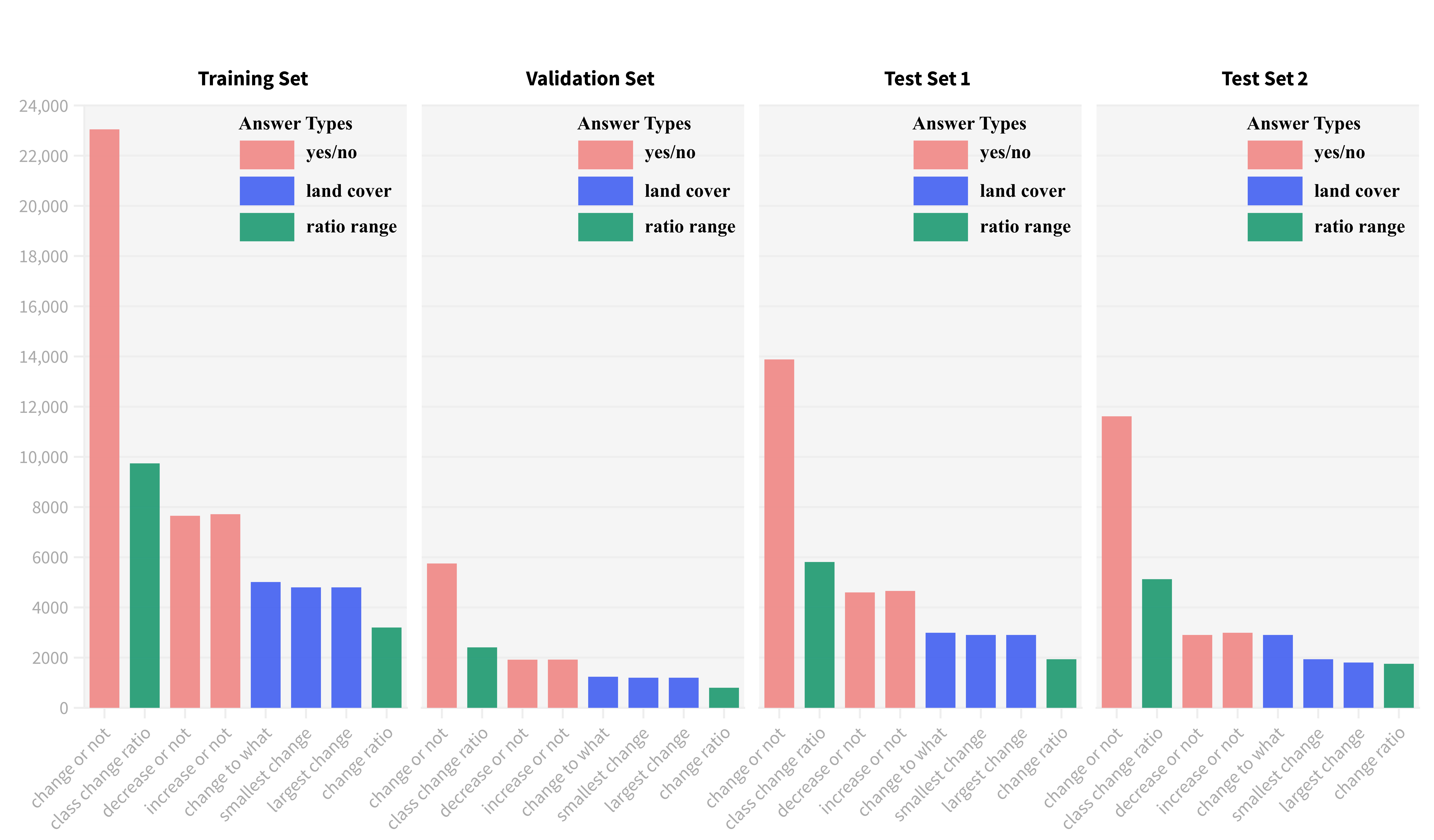}
	\caption{Visualization of question distributions of different subsets. From left to right: training set, validation set, test set 1, and test set 2.}
	\label{CV3}
\end{figure}

\item \textbf{Change to what}. 

This type of questions involves more detailed information about changes, i.e., what the land-cover at time $\text{T}_1$ mainly becomes to at time $\text{T}_2$. Such questions require analyzing the same region in multi-temporal images to obtain the change of land-cover types in this region. Although one class may change to more than one class over time, it is more meaningful to focus on the major change. Particularly, for a semantic class, we first find its pixel indices in ${\bm{s}_{t_1}}$. Then, the indices are used to select the corresponding pixels in ${\bm{s}_{t_2}}$. Finally, we count the number of the selected pixels for each land-cover type and choose the type with the largest number as the major change. In this case, the answer to the question what the regions of $l_i$ at time $\text{T}_1$ mainly change to should be the major change type.

\item \textbf{Largest/smallest change}. 

\emph {Largest/smallest change for an image pair}. Such questions focus on the largest or smallest changes in multi-temporal images. For each land-cover type, all changes in the two images should be taken into account. Therefore, the changed area for the land-cover class $l_i$ is ${\bm{A}_{t_1}^{i}}+{\bm{A}_{t_2}^{i}}$. By traversing all change types, the maximum and minimum changed regions can be obtained, and the corresponding land-cover classes are answers to this type of questions. In this dataset, the smallest change is which has the smallest changed area, and the unchanged type is not considered. 

\emph {Largest/smallest change for a single image}. To extract more detailed information about changes, we also analyze the maximum and minimum changed regions for the pre-event and post-event image separately. The maximum and minimum changed regions at time $\text{T}_1$ can be easily obtained by ${\arg\max_{l_i}(\bm{A}_{t_1}^{i})}$ and ${\arg\min_{li}}({\bm{A}_{t_1}^{i}})$, and the selected land-cover $l_i$ is the corresponding answer. For time $\text{T}_2$, the generation process is the same. This type of questions requires a model to not only identify land-cover changes in bi-temporal images but also understand which image ($\text{T}_1$ or $\text{T}_2$) is queried by users. In this context, the question “What is the smallest change in the first image?” is actually asking about the land-cover of the smallest changed region in the image captured at an earlier date.

\item \textbf{Change ratio}. 

\emph{Change ratio for all land-covers}. The percentages of changed regions are also very important information in practical applications. The change ratio can be calculated via dividing the changed area by the total area of the whole map, and the same for non-change ratio.
Since proportions are continuous numbers, they cannot be compatible with the classification task. Thus, we discretize ratios into bins. 
To be more specific, numerical answers are quantized into 11 categories: 0\%, 0\%-10\%, 10\%-20\%, 20\%-30\%, 30\%-40\%, 40\%-50\%, 50\%-60\%, 60\%-70\%, 70\%-80\%, 80\%-90\%, and 90\%-100\%. Notice that in this context A\%-B\% means $(A,B]$. In this way, we calculate the change percentage for each image pair and gain answers to the change ratio-related questions.

\emph{Change ratio for each land-cover}. In addition to the ratio of all changed regions, we also want to analyze the change ratio for each land-cover class on the pre-event or post-event image. Similarly, numerical answers are also quantized as above. For each land-cover class $l_i$, we first calculate its changed regions ${\bm{A}_{t_1}^{i}}$ and ${\bm{A}_{t_2}^{i}}$ at $\text{T}_1$ and $\text{T}_2$. Then, the change ratio for $l_i$ on the pre-event image is calculated via dividing ${\bm{A}_{t_1}^{i}}$ by the total area of the whole image. In the same way, change ratios for different land-covers on the post-event image can be obtained.
\end{itemize}

\begin{figure*}
	\centering
	\includegraphics[width=1.0\textwidth]{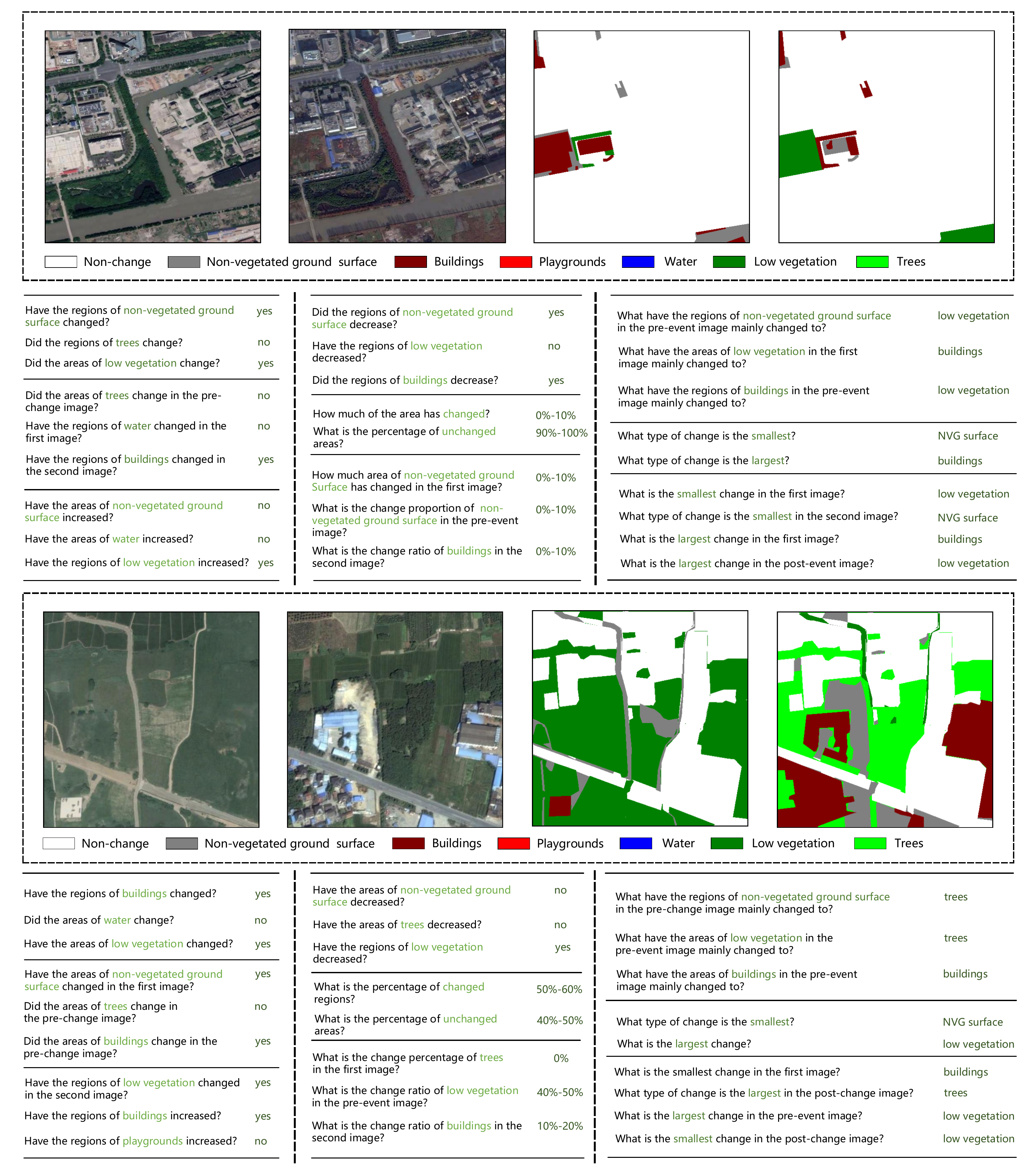}
	\caption{Visualization examples of the generated CDVQA dataset. Here, we show two data samples, and each one contains bi-temporal images, questions, and the corresponding answers. Best viewed in color and zoomed in.}
	\label{mainarch1}
\end{figure*}

In practice, we have defined multiple synonymous templates for each type of questions. During the question-answer generation process, for each image pair, question-answer pairs are generated separately for each question type. As more than one template is designed for each question type, we randomly select one of them to generate a sample. To balance the number of samples in each question type, we set different probabilities for generating samples of different question types. Specifically, we set a low probability value for the “yes/no” type and a high probability value for other question types. For each image pair, we generate 16 samples in average.

\subsection{Question and Answer Distributions}
As 2,968 pairs of images are publicly available, we use these images as the basic data to generate the CDVQA dataset. The whole dataset is split into the training set, validation set, and test set. To better evaluate the robustness and reliability of CDVQA models, we generate two test sets with different distributions of answers. The class distributions of answers in the generated CDVQA dataset are displayed in Fig. \ref{answer_type_distribution}. From this figure, we can see that the training set, validation set, and test set 1 share the same class distribution. The answer distributions of test set 1 and test set 2 are different.

As we can see from Fig. \ref{answer_type_distribution}, answer types in all subsets obey the long-tail distribution. Concretely, answer class \emph{no} dominates answer distributions in all subsets. For example, in the training set, samples with answer \emph{no} occupy 30.9\% of all instances. In the test set 1, answers \emph{no} account for 31.15\% of total answers. In contrast, answers \emph{50\%-60\%} only occupy 0.22\% of all answers. The reason for the class imbalance is that there are more questions asking for \emph{yes} or \emph{no}. The answers to questions such as \emph{change or not} and \emph{increase/decrease or not} are \emph{yes} or \emph{no}. 

The question type distributions of all four subsets are presented in Fig. \ref{CV3}. For simplicity, \emph{change ratio for each land-cover} is denoted as \emph{class change ratio}. We can see that distributions of question types are also long-tailed. In addition, question type \emph{change or not} has the highest frequencies in all subsets. This is the reason why the two most frequent answer types are \emph{yes} and \emph{no}. Similar to answer distributions, the distributions of question types for the training set, validation set, and test set 1 are the same, while they are different from the distribution of the test set 2. Specifically, the proportions of questions about “change to what”, “change ratio”, and “class change ratio” increase in test set 2 compared to test set 1. Since questions of these types are more difficult, test set 2 is more difficult than test set 1. Visualization examples of the generated CDVQA dataset are shown Fig. \ref{mainarch1}.
 
\section{Methodology}
\label{Methodology}

In this work, the CDVQA task is deemed as a classification task. Note that semantic change maps are only used to generate question-answer pairs in the dataset preparation phase, and in CDVQA, only image pairs, questions, and the corresponding answers are used for training and evaluating a model. As shown in Fig. \ref{mainarch}, our CDVQA model takes as input two aerial images and a question. The output of the model is an answer predicted by the network. Particularly, the whole network architecture consists of four parts. The first component is a multi-temporal visual feature learning module which is used to encode the input images into deep features. The second part, named multi-temporal fusion, is responsible for fusing the features of the two images. The third one is a multi-modal fusion module that aims at fusing the image and question features. The forth is an answer prediction part, which takes the fused multi-modal feature as input to predict the answer.
In addition, for CDVQA task, we design a change enhancing module to encourage the model to focus on changed pixels of the input images.
The proposed modules in our CDVQA framework will be described in detail in the following subsections.

\subsection{Multi-temporal Encoder}
Different from tasks like image classification, object detection, and semantic segmentation, change analysis involves two input images of the same location but at different times. Similarly, a CDVQA system takes as input multi-temporal inputs.
In order to identify changes between two images, temporal differences should be extracted and analyzed. 

In respect of multiple inputs, Siamese networks are commonly used in many vision tasks. We denote the feature of the image of time $\text{T}_1$ as $\bm{F}_1=f_{1}(\bm{x}_{t_1})$. Likewise, $f_2(\cdot)$ is used to obtain the encoded representation for the image of time $\text{T}_2$. For Siamese networks, we set the network architecture and parameters of $f_1$ and $f_2$ to be the same.

In this work, we explore effects of different encoder networks on CDVQA. For visual feature extraction, convolutional neural networks (CNNs) are usually used to learn feature representations, and ResNet \cite{he2016deep} is an important milestone in the development of CNN architectures. Thus, different scales of ResNets, e.g., ResNet-18, ResNet-101, and ResNet-152, are employed as the multi-temporal encoder of our CDVQA model, aiming at studying effects of different scales of CNNs on CDVQA.

Recently, Transformer architecture \cite{vaswani2017attention} has achieved excellent performance on NLP tasks \cite{tetko2020state}. Designed for sequence modeling tasks, Transformer has the significant advantage of using attention to learn long-range dependencies in data. Considering its great success in the language modeling domain, it has also been applied to computer vision tasks, to name a few, image classification \cite{liu2021swin,dosovitskiy2020image}, object detection \cite{carion2020end}, and semantic segmentation \cite{xie2021segformer}. In this work, the Transformer-based encoder for multi-temporal images is also used. 


\subsection{Change Enhancing Module}
Change detection is a fundamental task in remote sensing and also the core of CDVQA task. To answer change-related questions, a model needs to focus on changed regions and further analyze semantic information. In a number of computer vision tasks, self-attention mechanism \cite{xiong2020msn, liu2021dual, xiong2021ask} is used to boost the performance by focusing on important parts of data samples. However, there are two input images in our case, where the self-attention mechanism is not applicable. Hence, in this work, we propose a change enhancing module to enhance the CDVQA model in terms of the capability of detecting changes.

We denote that the encoded deep features for the input two images are $\bm{F}_1\in \R^{N\times C\times H\times W}$ and $\bm{F}_2\in \R^{N\times C\times H\times W}$, respectively, where $N$ is batch size, $C$ is the number of channels, and $H$, $W$ are the height and width of feature maps. The conventional self-attention model \cite{vaswani2017attention} first transforms the input feature into three independent features, i.e., the query $\bm{Q}$, key $\bm{K}$, and value $\bm{V}$. In contrast, for the proposed change enhancing module, we treat the feature representations $\bm{F}_1$ and $\bm{F}_2$ as the query and key, respectively, and compute their similarity $\bm{F}_s \in \R^{N\times C\times H\times W}$ as follows:
\begin{equation}
	\bm{F}_s = |f_q(\bm{F}_1) - f_k(\bm{F}_2)|,
\end{equation}
where $f_q(\cdot)$ and $f_k(\cdot)$ are $1\times 1$ convolutions for the purpose of feature transformation. Next, a change enhancing map $\bm{M}_{ce} \in \R^{N\times H\times W}$ can be obtained by:
\begin{equation}
\bm{M}_{ce} = \sigma(f_c(\bm{F}_s)),
\end{equation}
where $f_c(\cdot)$ is a $1\times 1$ convolution layer for predicting the change enhancing map. $\sigma(\cdot)$ is ReLU activation function. The map $\bm{M}_{ce}$ is used to encourage the model to focus on regions where differences between $\bm{F}_1$ and $\bm{F}_2$ are large. To this end, we scale $\bm{M}_{ce}$ with a parameter $\theta$ and add it with an identity matrix $\bm{I}$. $\theta$ is a learnable parameter with an initial value of 0 and is optimized during training in an end-to-end manner. Then, we multiply the transformed $\bm{M}_{ce}$ 
and two encoded features, respectively:
\begin{equation}
\begin{split}
\bm{F}_{c1} = (\bm{I} + \theta \bm{M}_{ce})  \bm{F}_1, \\
\bm{F}_{c2} = (\bm{I} + \theta \bm{M}_{ce})  \bm{F}_2,
\end{split}
\end{equation}
where $\bm{F}_{c1}$ and $\bm{F}_{c2}$ are the final encoded features corresponding to two input images. $\bm{F}_{c1}$ and $\bm{F}_{c2}$ will then be fused by the multi-temporal feature fusion module, which will be introduced in the next subsection.

\subsection{Multi-temporal Fusion}
After the feature encoding and change enhancing processing, we need to fuse features of time $\text{T}_1$ and time $\text{T}_2$ to obtain the final visual feature $\bm{F}_v$. For the fusion of multiple feature maps, element-wise subtraction, multiplication, summation, and concatenation are commonly used methods.
Given two feature maps $\bm{F}_{c1} \in \R^{N\times C\times H\times W}$ and $\bm{F}_{c2} \in \R^{N\times C\times H\times W}$, the aforementioned fusion methods can be formulated as:
\begin{equation}
\begin{split}
 	\bm{F}_{v1}&=\bm{F}_{c1} \ominus \bm{F}_{c2}, \\
 	\bm{F}_{v2}&=\frac{\bm{F}_{c1}}{{||\bm{F}_{c1}||}_2}\ominus \frac{\bm{F}_{c2}}{{||\bm{F}_{c2}||}_2},\\
 	\bm{F}_{v3}&=\bm{F}_{c1}\,^ \frown \bm{F}_{c2}, \\
 	\bm{F}_{v4}&=\bm{F}_{c1} {\oplus} \bm{F}_{c2}, \\
 	\bm{F}_{v5}&=\bm{F}_{c1} \otimes \bm{F}_{c2},
 \end{split}
\end{equation}
where $\ominus$ denotes element-wise subtraction operation. Note that we normalize the two features before the element-wise subtraction operation for computing $\bm{F}_{v2}$. $\oplus$ and $\otimes$ denote element-wise summation and multiplication operations, respectively. $^\frown$ stands for the concatenation operation along the channel dimension. To study effects of different fusion strategies, we compare and analyze their performance in Section \ref{Experiments}.

\subsection{Multi-modal Fusion}
Since CDVQA involves both visual features and language representations, we need to fuse multi-modal features. After the multi-temporal feature fusion, the final visual representation $\bm{F}_v \in \R^{N\times C\times H\times W}$ can be obtained. Meanwhile, a recurrent neural network (RNN) is used to encode the question into feature vector $\bm{V}_q \in \R^{N\times L}$. As the skip-thoughts model has been applied in many remote sensing image-based NLP tasks \cite{lobry2020rs, chappuis2021find}, we choose to use the pre-trained skip-thoughts model \cite{kiros2015skip} for the language feature extraction part. Specifically, skip-thought vectors are modeled with an encoder-decoder architecture, and both are constructed with RNNs. The encoder transforms the input sentence into a vector, and two decoders are used to decode the vector into the previous and the next sentence, respectively. In this work, we use the encoder of skip-thoughts for generating language embeddings.

Before fusing features of two modalities, we first transform the visual feature $\bm{F}_v$ into a feature vector $\bm{F}_{vt} \in \R^{N\times L}$. Then, the two feature vectors have the same size, and we can fuse $\bm{F}_{vt}$ and $\bm{V}_q$ together. As how to fuse them is not the main research content of this work, we simply merge them into a multi-modal feature by concatenation:
\begin{equation}
	\begin{split}
		\bm{F}_m=\bm{V}_q \,^ \frown \bm{F}_{vt}, \\
	\end{split}
\end{equation}
where $\bm{F}_m\in \R^{N\times 2L}$ is the fused multi-modal representation. 

Finally, as the answer prediction is modeled as a classification task in this work. The feature $\bm{F}_m$ is used to predict the answer by passing through a classifier, i.e., two fully connected layers. The answer is given by selecting the answer class with the highest probability. The output dimension of the first layer is 256 and the final output dimension of the classifier is 19, as there are 19 answer types. Specifically, the possible answers include no, yes, 0\%-10\%, 0, NVG surface, buildings, low vegetation, 10\%-20\%, trees, 20\%-30\%, water, 80\%-90\%, 30\%-40\%, 90\%-100\%, 70\%-80\%, 40\%-50\%, 60\%-70\%, 50\%-60\%, and playgrounds (sorted by the number of samples).

\section{Experiments}
\label{Experiments}
\begin{table}[]
	\centering
	\caption{Numerical Results of Using Different Backbone Networks on the Test set 1 of CDVQA Dataset.}
	\label{tabel:l1}
	\scalebox{0.94}{
		\begin{tabular}{ccccc}
		  \toprule[1pt]
			Question Types     & ResNet-18       & ResNet-101      & ResNet-152      & ViT-B16         \\  \midrule[1pt]
			change ratio       & 0.3455          & 0.3388          & 0.3476          & \textbf{0.3600} \\
			class change ratio  & 0.7200          & 0.7134          & 0.7115          & \textbf{0.7231} \\
			change or not      & 0.8379          & 0.8387          & 0.8374          & \textbf{0.8401} \\
			change to what     & 0.5710          & 0.5737          & 0.5770          & \textbf{0.5820} \\
			increase or not    & 0.6913          & 0.6902          & 0.6854          & \textbf{0.6952} \\
			decrease or not    & \textbf{0.7303} & 0.7243          & 0.7275          & 0.7185          \\
			smallest change    & 0.2627          & \textbf{0.2758} & 0.2734          & 0.2710          \\
			largest change     & 0.4603          & 0.4576          & \textbf{0.4669} & 0.4648          \\ \midrule[0.6pt]
			Average Accuracy   & 0.5773          & 0.5766          & 0.5783          & \textbf{0.5818} \\
			Overall Accuracy   & 0.6771          & 0.6763          & 0.6766          & \textbf{0.6800} \\ \bottomrule[1pt]
	\end{tabular}}
\end{table}

\begin{table}[t]
	\centering
	\caption{Numerical Results of Using Different Backbone Networks on the Test set 2 of CDVQA Dataset.}
	\label{tabel:l2}
	\scalebox{0.94}{
		\begin{tabular}{ccccc}
		\toprule[1pt]
			Question Types     & ResNet-18 & ResNet-101      & ResNet-152      & ViT-B16         \\ \midrule[1pt]
			change ratio       & 0.3444    & 0.3465          & 0.3588          & \textbf{0.3651} \\
			class change ratio & 0.7131    & 0.7158          & 0.7142          & \textbf{0.7174} \\
			change or not      & 0.7818    & 0.8363          & \textbf{0.8403} & 0.8353          \\
			change to what     & 0.5316    & 0.5693          & 0.5705          & \textbf{0.5790} \\
			increase or not    & 0.7003    & 0.6880          & \textbf{0.7012} & 0.6847          \\
			decrease or not    & 0.7142    & \textbf{0.7331} & 0.7264          & 0.7303          \\
			smallest change    & 0.2000    & \textbf{0.2803} & 0.2568          & 0.2592          \\
			largest change     & 0.3793    & 0.4607          & 0.4637          & \textbf{0.4665} \\ \midrule[0.6pt]
			Average Accuracy   & 0.5456    & 0.5787          & 0.5790          & \textbf{0.5797} \\
			Overall Accuracy   & 0.6165    & 0.6333          & 0.6336          & \textbf{0.6341} \\ \bottomrule[1pt]
	\end{tabular}}
\end{table}

\begin{figure}[t]
	\centering
	\includegraphics[width=0.46\textwidth]{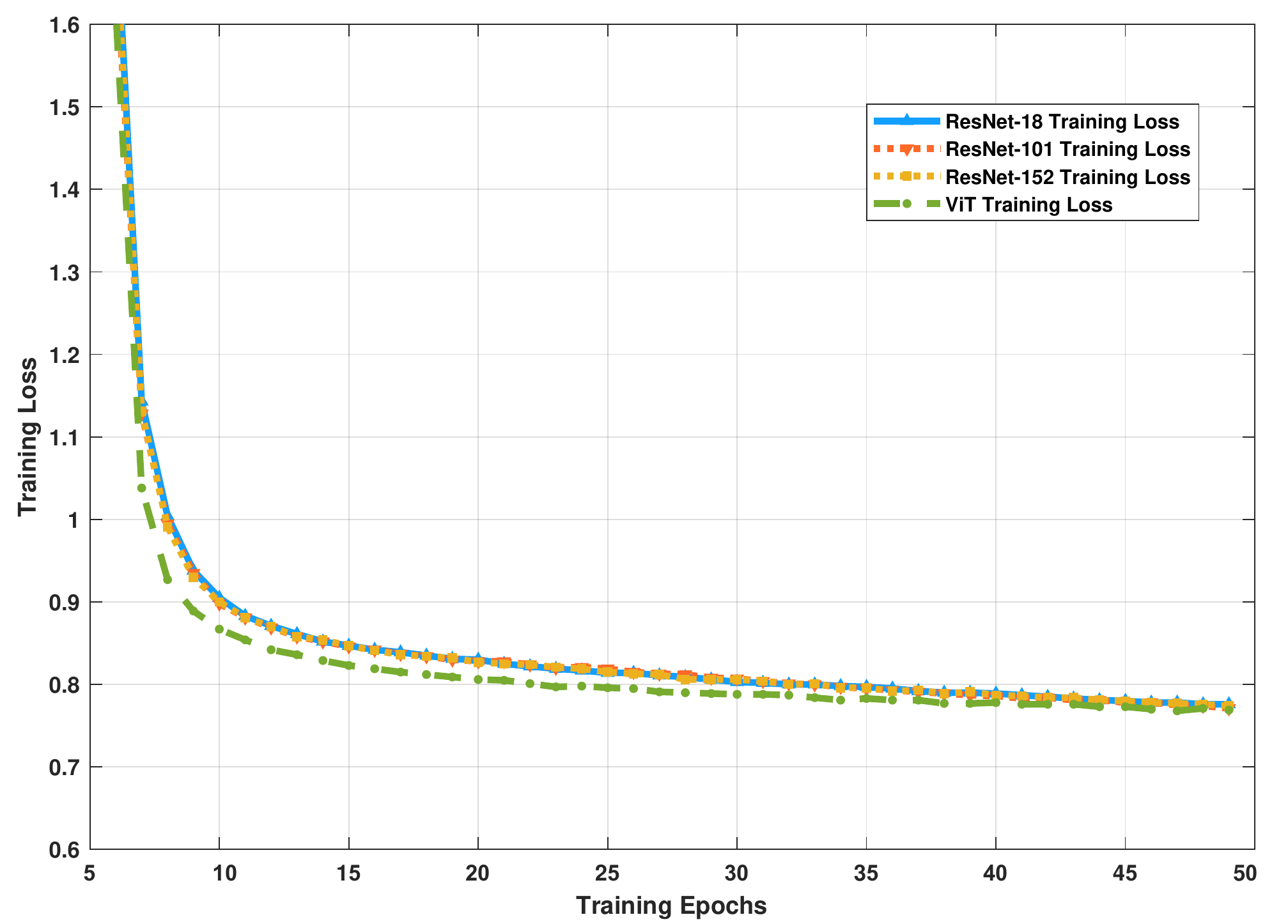}
	\caption{Visualization of training losses. Four different backbone networks are compared. }
	\label{Trainloss}
\end{figure}

\begin{figure}
	\centering
	\includegraphics[width=0.46\textwidth]{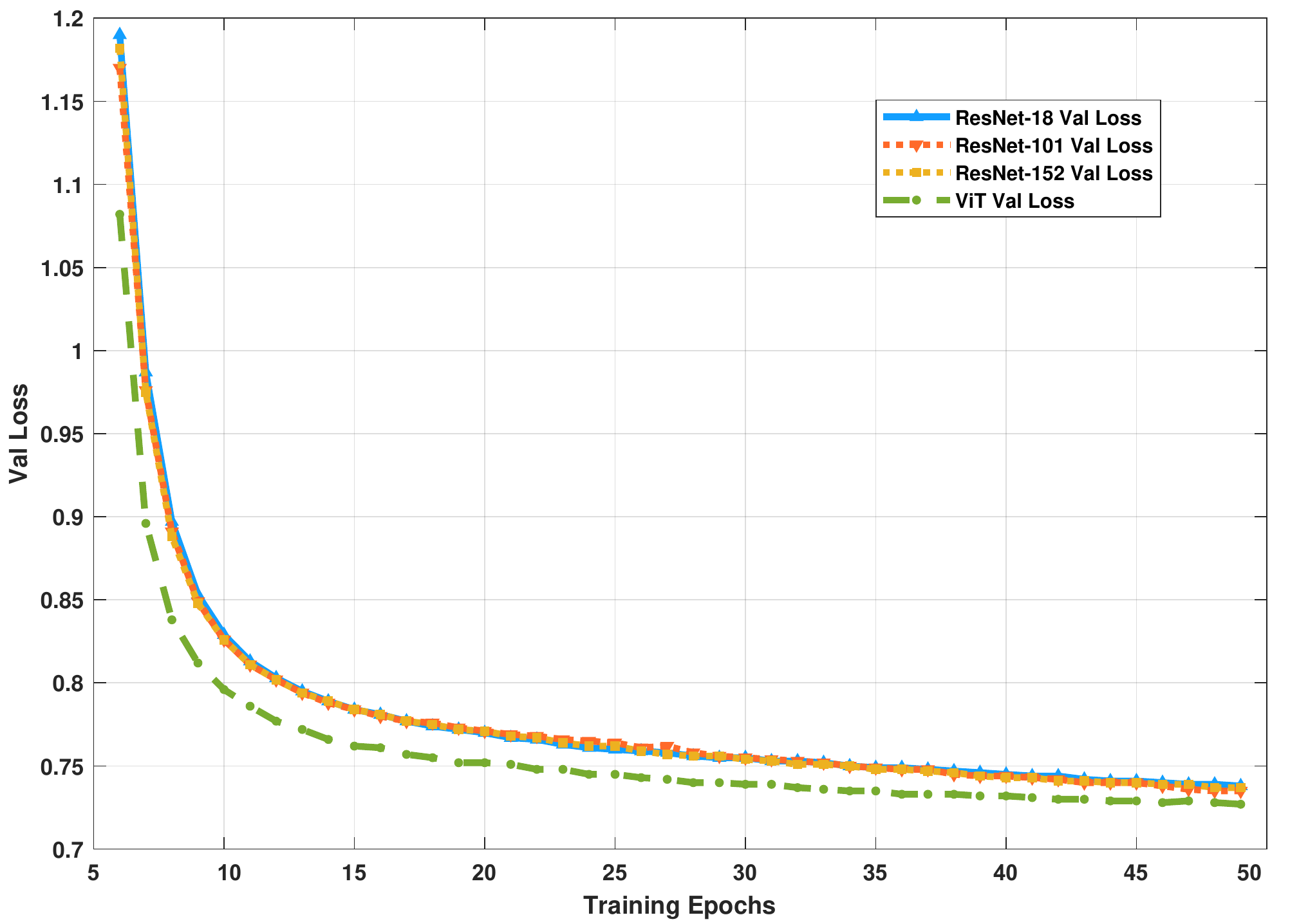}
	\caption{Visualization of validation losses. Four different backbone networks are compared. }
	\label{ValLoss}
\end{figure}

\subsection{Datasets}

The CDVQA dataset is publicly available in 2,968 image pairs with the size of $512 \times 512$. Based on these image pairs, there are more than 122,000 question-answer pairs generated in total. The training, validation, and test sets are split based on image pairs captured at different geographical positions. Particularly, the training set contains 65,967 question-answer pairs, which are generated from 1,600 (53.91\%) image pairs. There are 16,441 question-answer pairs in the validation set, which are produced based on 400 (13.48\%) image pairs. Besides, we use the left 968 (32.61\%) image pairs to generate two test sets with 39,686 (test set 1) and 31,036 (test set 2) question-answer pairs for more comprehensive model evaluation. Note that there is an overlap between the two test sets.
\begin{figure*}
	\centering
	\includegraphics[width=1.0\textwidth]{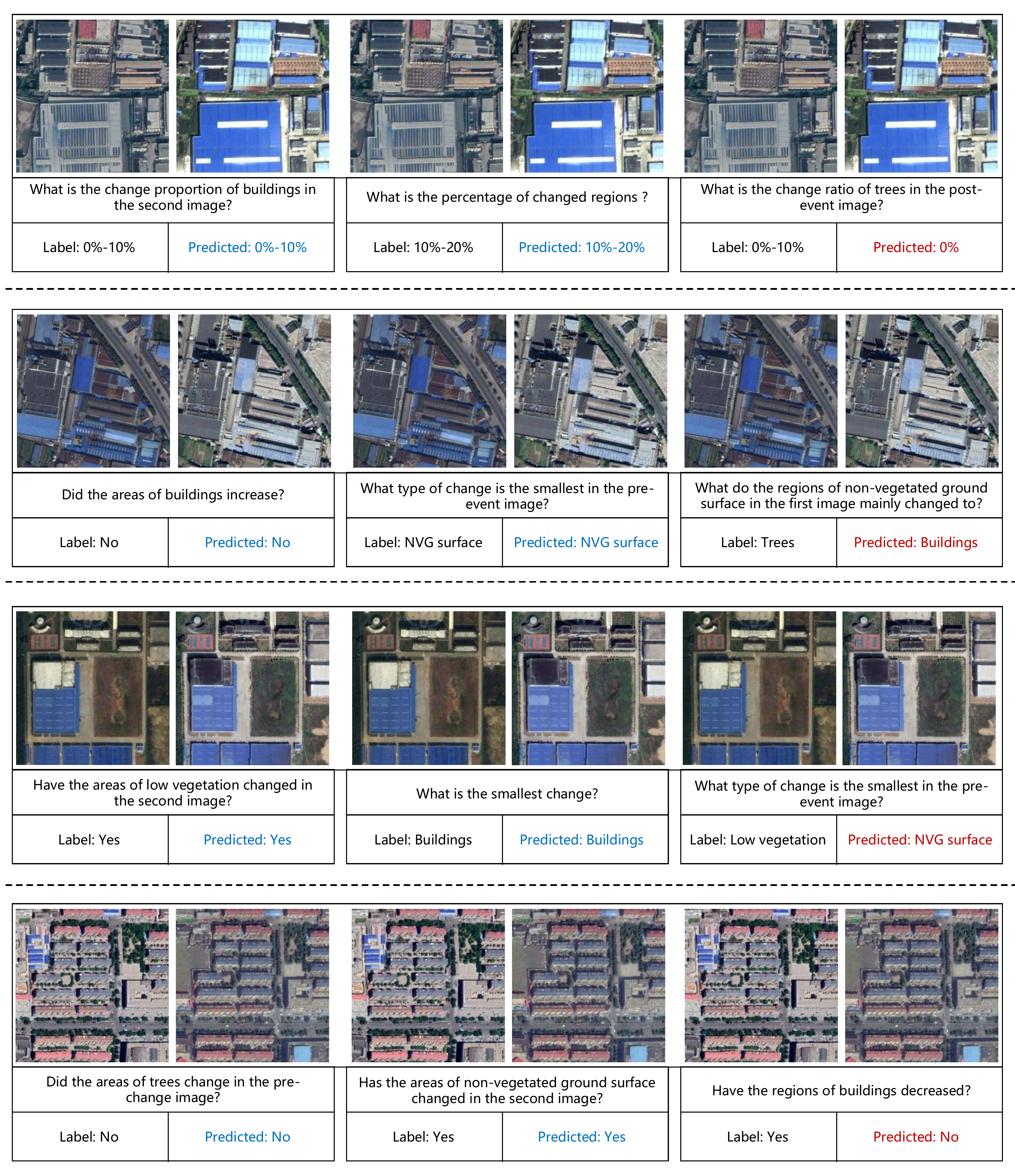}
	\caption{Visualization examples of CDVQA results. Each row presents three different questions and the same input image pair. Correctly predicted results are shown in \textcolor{blue1} {blue}, and wrong answers are in \textcolor{red} {red}.}
	\label{arch}
\end{figure*}

\subsection{Implementation Details}
The generated dataset for CDVQA follows the same format as the work of RSVQA \cite{lobry2020rs}. Regarding training parameters, Adam optimizer is used with an initial learning rate of \mbox{1e-4}. For all ResNet-based models, the batch size is set to 70, and the size of the input image is scaled to $256\times 256$. Since the used ViT \cite{dosovitskiy2020image} model requires the input size to be $384\times 384$, we have to reduce the batch size to 32 considering GPU memory limit. For all experiments, 50 epochs are used to train models. We utilize accuracy as a measurement for each question type. Additionally, average accuracy and overall accuracy are also reported.

\subsection{Effects of Different Backbones}
The backbone network of the visual encoder is an important component. Therefore, we compare four different backbones: three ResNets (ResNet-18, ResNet-101, and ResNet-152) and a vision Transformer model ViT. In all experiments, we fuse multi-temporal visual features by feature concatenation for all backbone networks. 

The results on two different test sets are displayed in Table \ref{tabel:l1} and Table \ref{tabel:l2}, respectively. From the results, we can see that compared to ResNet-18 and ResNet-101, ResNet-152 does not show a significant performance advantage.
For example, on the test set 1, ResNet-18 and ResNet-152 deliver very close average and overall accuracies. This indicates that merely improving the capability of backbone network for visual learning only yields a limited gain. However, when we change the network architecture of the backbone from ResNet to Transformer, the performance can be further improved. The reason for this improvement is that the self-attention mechanism of Transformer networks is beneficial for learning more representative features. Note that parameters of backbone networks are fixed during the training stage. In Fig. \ref{Trainloss} and Fig. \ref{ValLoss}, we also visualize training and validation losses of models with different backbones. It can be seen that the ViT backbone has significantly lower losses than ResNet-based networks. Note that we omit the first 5 epochs to better compare the final convergence state.

From the results, we can see that different backbone networks have very little impact on the performance of our framework. This is because visual feature learning may not be the key to improving accuracy. Other parts of the model, such as multi-temporal fusion and change analysis part, may be more critical for the performance improvement of the CDVQA task.

\begin{table}
	\centering
	\caption{Numerical Results of Using Different Fusion Strategies on the Test Set 1 of CDVQA Dataset.}
	\label{tabel:l3}
	\scalebox{1.0}{
		\begin{tabular}{m{2.1cm}<{\centering} m{0.7cm}<{\centering} m{0.7cm}<{\centering} m{0.7cm}<{\centering} m{0.7cm}<{\centering} m{0.7cm}<{\centering}}
			\toprule[1pt]
			ResNet-101 & Concat {$\frown$}   & Sub {$\ominus$} & NSub {$\ominus$} & Mul {$\otimes$} & Sum {$\oplus$} \\ \midrule[1pt]
			change ratio          & \textbf{0.3388} & 0.3182       & 0.3151             & 0.3047          & 0.3352    \\
			class change ratio    & 0.7134          & 0.7129       & 0.7130             & 0.7131          & \textbf{0.7167}    \\
			change or not         & \textbf{0.8387} & 0.8249       & 0.8245             & 0.8271          & 0.8378    \\
			change to what        & \textbf{0.5737} & 0.5269       & 0.5543             & 0.5693          & 0.5690    \\
			increase or not       & 0.6902          & 0.6897       & 0.6919             & \textbf{0.6967} & 0.6891    \\
			decrease or not       & \textbf{0.7243} & 0.6983       & 0.7056             & 0.7020          & 0.7239    \\
			smallest change       & 0.2758          & 0.2778       & 0.2789             & \textbf{0.2799} & 0.2703    \\
			largest change        & \textbf{0.4576} & 0.4473       & 0.4411             & 0.4290          & 0.4476    \\ \midrule[0.6pt]
			Average Accuracy      & \textbf{0.5766} & 0.5620       & 0.5655             & 0.5652          & 0.5737    \\
			Overall Accuracy      & \textbf{0.6763} & 0.6631       & 0.6657             & 0.6665          & 0.6746    \\
			\bottomrule[1pt]
		\end{tabular}
	}
\end{table}

\begin{table}
	\centering
	\caption{Numerical Results of Using Different Backbone Networks on the Test Set 2 of CDVQA Dataset.}
	\label{tabel:l4}
	\scalebox{1.0}{
		\begin{tabular}{m{2.1cm}<{\centering} m{0.7cm}<{\centering} m{0.7cm}<{\centering} m{0.7cm}<{\centering} m{0.7cm}<{\centering} m{0.7cm}<{\centering}}
			\toprule[1pt]
			ResNet-101         & Concat {$\frown$}  & Sub {$\ominus$} & NSub {$\ominus$} & Mul {$\otimes$} & Sum {$\oplus$}      \\ \midrule[1pt]
			change ratio       & \textbf{0.3465} & 0.3223       & 0.3243             & 0.3119         & 0.3409          \\
			class change ratio & 0.7158          & 0.7097       & 0.7103             & 0.7104         & \textbf{0.7169} \\
			change or not      & \textbf{0.8363} & 0.8195       & 0.8228             & 0.8148         & 0.8347          \\
			change to what     & \textbf{0.5693} & 0.5275       & 0.5549             & 0.5396         & 0.5690          \\
			increase or not    & 0.6880          & 0.6986       & \textbf{0.7146}    & 0.7036         & 0.6886          \\
			decrease or not    & 0.7331 & 0.7029       & 0.7046             & 0.7172         & \textbf{0.7423}          \\
			smallest change    & \textbf{0.2803} & 0.2792       & 0.2796             & 0.2799         & 0.2679          \\
			largest change     & \textbf{0.4607} & 0.4442       & 0.4428             & 0.4290         & 0.4486          \\ \midrule[0.6pt]
			Average Accuracy   & \textbf{0.5787} & 0.5630       & 0.5692             & 0.5632         & 0.5761          \\
			Overall Accuracy   & \textbf{0.6333} & 0.6199       & 0.6244             & 0.6196         & 0.6313          \\ \bottomrule[1pt]
		\end{tabular}
	}
\end{table}

\subsection{Effects of Different Multi-temporal Fusion Strategies}
In this subsection, we quantitatively compare five commonly-used feature fusion operations, namely concatenation (Concat), summation (Sum), subtraction (Sub), normalized subtraction (NSub), and multiplication (Mul). The numerical results on two test sets are presented in Table \ref{tabel:l3} and Table \ref{tabel:l4}. The results in these tables show that concatenation is the best. The concatenation operation first concatenates two inputs together, and then several fully-connected layers are used to fuse these inputs by learnable weights. This makes it a more flexible and general fusion strategy. 

For change analysis tasks, intuitively, subtraction should be the best fusion method, as it can better highlight changed regions. However, it can be seen from the results that the subtraction operation cannot outperform others. Considering that the direct subtraction of two features may undermine their representability, we normalize two input features by using $\ell_2$ normalization before the subtraction operation. Nevertheless, the normalized subtraction operation is still no better than concatenation and summation. This indicates that directly subtracting two inputs is not useful to CDVQA tasks, and a specific change analysis module should be designed.

\subsection{Effect of Change Enhancing Module}
It is critical to obtain semantic change information from multi-temporal images. However, there are no pixel-wise ground truth change labels available in this task. To incorporate change information into the model, we propose a change enhancing module to highlight changed regions in the input images. To validate the effectiveness of the module, we conduct an ablation study, and numerical results are displayed in Table \ref{tabel:l5} and Table \ref{tabel:l6}. In the two tables, change enhancing module is abbreviated as CEM for the sake of simplification. The experimental results on both test sets indicate that the proposed change enhancing module is beneficial to the CDVQA task. In particular, from the results in two tables, it can be seen that the proposed module can consistently improve both the average accuracy and overall accuracy.

\subsection{Cross-dataset Evaluation}
In order to explore the generalization ability of the model, we construct another CDVQA dataset as an additional test set. Specifically, we collect 138 image pairs of size 256×256 from HTCD \cite{shao2021sunet} dataset (only binary change maps are available) and manually annotate semantic change maps. Then, 3,303 question-answer pairs are generated and used for the cross-dataset test setting. To show the effectiveness of the proposed method, we compare the performance of a model trained on the CDVQA dataset and another model with randomly initialized weights. Table \ref{tabel:l7} shows numerical results. We can see that the model trained on our CDVQA dataset can be transferred to unseen scenarios, but its performance is not satisfactory in this cross-dataset test setting. This is mainly because there is a domain gap between the tests set of CDVQA and this one. We see that much more research efforts are needed in this direction.

\subsection{Discussion}
\begin{figure}
	\centering
	\includegraphics[width=0.5\textwidth]{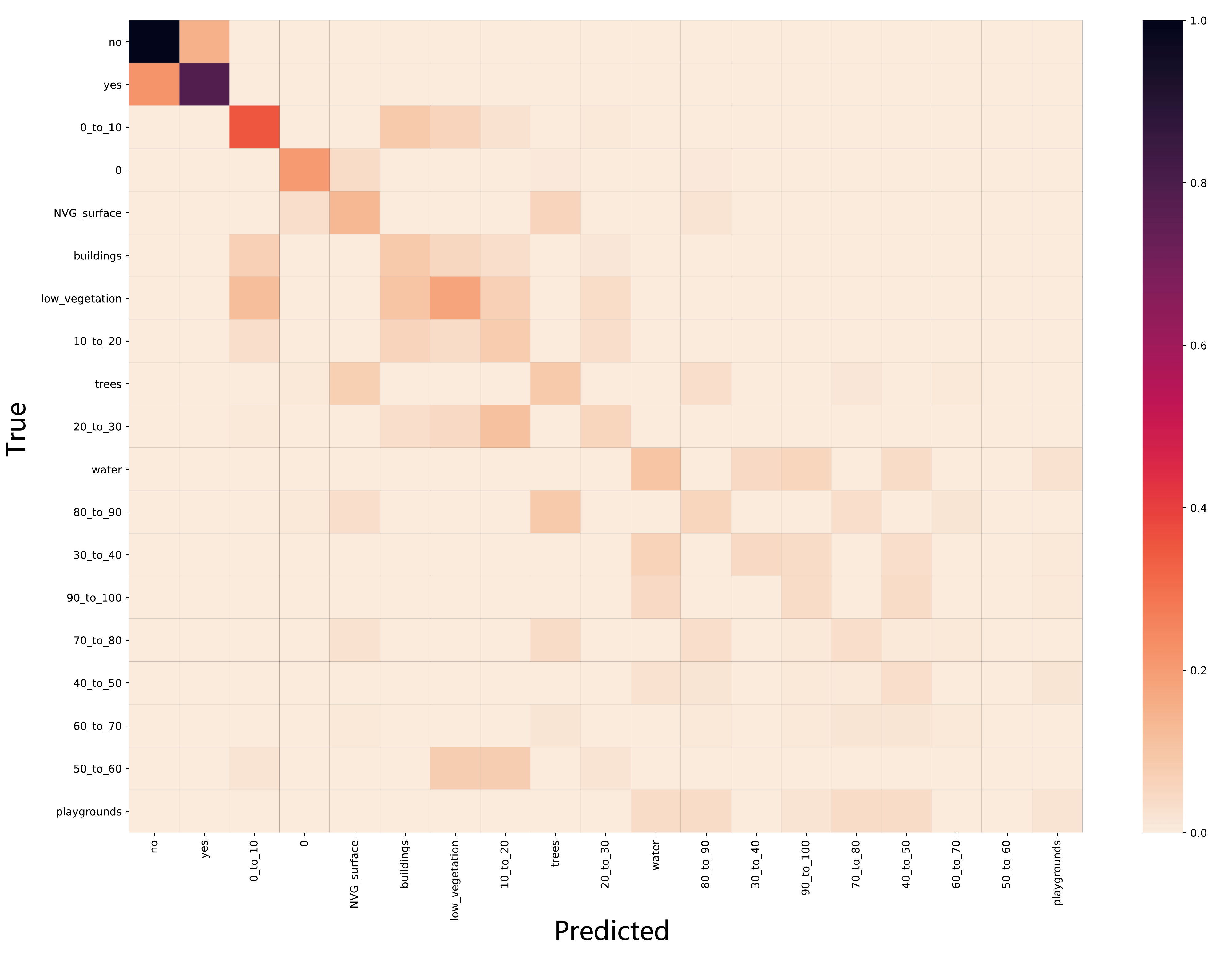}
	\caption{{Normalized confusion matrix for our CDVQA dataset on the test set 1 (ResNet-152 is used as the backbone).}}
	\label{cfm}
\end{figure}

Regarding numerical results, generally, the average accuracy is lower than 60\%, and the overall accuracy is lower than 70\%. Some visualization examples of CDVQA results are presented in Fig. \ref{arch}. Both correctly predicted examples and failures are displayed. From the experimental results, we can conclude that CDVQA is a complex and challenging task. To correctly answer different types of questions, a model first needs to learn multi-modal representations for the input images and questions. Visual and language understanding is of great importance for the model. Besides, CDVQA also requires the model to be able to analyze semantic change information. I.e., the model needs to not only locate changed areas, but also identify land-cover classes of changed regions to answer some complex questions. Currently, the proposed baseline framework does not make use of semantic change labels. Thus its performance on questions related to land-cover classes is not that satisfactory. Change ratio for each land-cover has higher accuracy than change ratio for all land-covers. This is mainly because the former has more training samples. We also visualize the normalized confusion matrix in Fig. \ref{cfm}. Note that the confusion matrix is normalized along the predicted label axis.
 
\begin{figure*}
	\centering
	\includegraphics[width=1.0\textwidth]{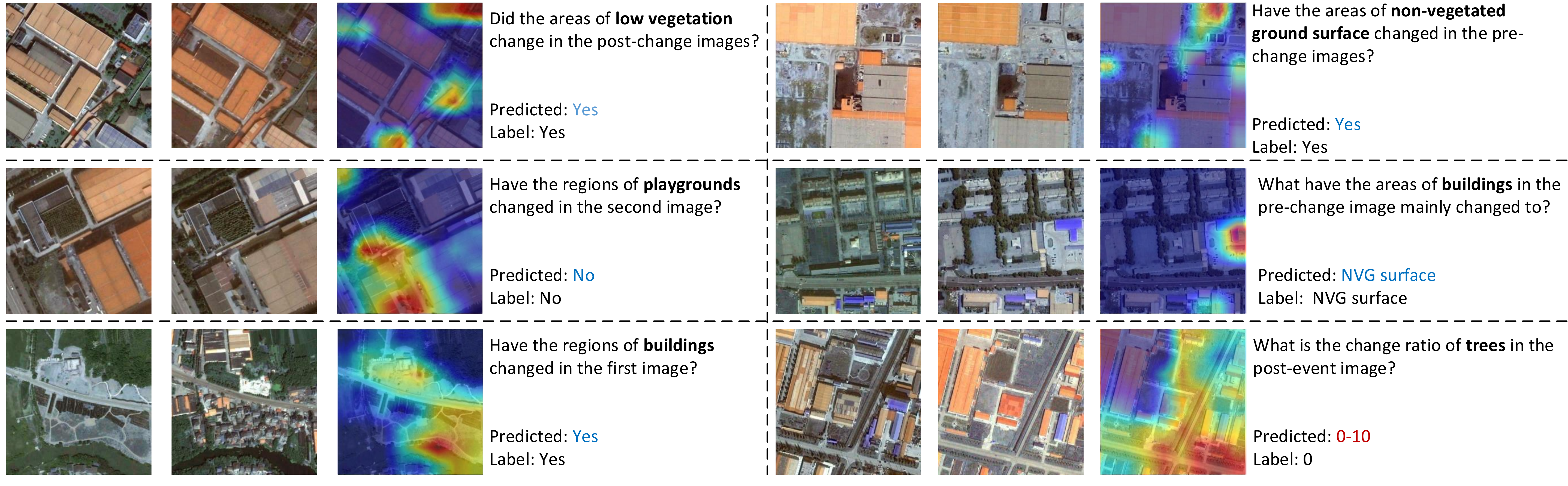}
	\caption{Visualization examples of change attention maps. Best viewed in color.}
	\label{att_vis}
\end{figure*}

\begin{table}[]
	\centering
	\caption{{Ablation Study on the Test Set 1 of CDVQA Dataset for ResNet-101 backbone.}}
	\label{tabel:l5}
	\scalebox{1.08}{
	\begin{tabular}{m{2.1cm}<{\centering} m{1.6cm}<{\centering} m{1.6cm}<{\centering}}
		\toprule[1pt]
		Question Types     & w/o CEM      & w/ CEM  \\ \midrule[1pt]
		change ratio       & 0.3388          & \textbf{0.3854} \\
		class change ratio & \textbf{0.7134}          & 0.7071 \\
		change or not      & \textbf{0.8387}          & 0.8292 \\
		change to what     & 0.5737          & \textbf{0.5804} \\
		increase or not    & 0.6902          & \textbf{0.7443} \\
		decrease or not    & 0.7243          & \textbf{0.7697} \\
		smallest change    & 0.2758          & \textbf{0.3127} \\
		largest change     & 0.4576          & \textbf{0.4777} \\ \midrule[0.6pt]
		Average Accuracy   & 0.5766          & \textbf{0.6008} \\
		Overall Accuracy   & 0.6763          & \textbf{0.6903} \\ \bottomrule[1pt]
	\end{tabular}}
\end{table}

\begin{table}[]
	\centering
	\caption{Ablation Study on the Test Set 2 of CDVQA Dataset for ResNet-101 backbone.}
	\label{tabel:l6}
	\scalebox{1.1}{
		\begin{tabular}{m{2.1cm}<{\centering} m{1.6cm}<{\centering} m{1.6cm}<{\centering}}
			\toprule[1pt]
			Question Types     & w/o CEM      & w/ CEM  \\ \midrule[1pt]
			change ratio       & 0.3465          & \textbf{0.3904} \\
			class change ratio & 0.7158          & \textbf{0.7714} \\
			change or not      & \textbf{0.8363} & 0.8123             \\
			change to what     & 0.5693          & \textbf{0.6511} \\
			increase or not    & 0.6880          & \textbf{0.7534} \\
			decrease or not    & 0.7331          & \textbf{0.7868} \\
			smallest change    & \textbf{0.2803} & 0.2431          \\
			largest change     & \textbf{0.4607} & 0.3766          \\ \midrule[0.6pt]
			Average Accuracy   & 0.5787          & \textbf{0.5982} \\
			Overall Accuracy   & 0.6333          & \textbf{0.6513} \\ \bottomrule[1pt]
	\end{tabular}}
\end{table} 

\begin{table}[]
	\centering
	\caption{Experimental Results in the Cross-dataset Test Setting.}
	\label{tabel:l7}
	\scalebox{1.1}{
		\begin{tabular}{m{2.1cm}<{\centering} m{1.6cm}<{\centering} m{1.6cm}<{\centering}}
			\toprule[1pt]
			Question Types     & Random init.      & Ours  \\ \midrule[1pt]
			change ratio       & 0.0442          & \textbf{0.0615} \\
			class change ratio & 0.0362          & \textbf{0.0751} \\
			change or not      & {0.0730} & \textbf{0.2042}             \\
			change to what     & 0.0722          & \textbf{0.0834} \\
			increase or not    & 0.0382          & \textbf{0.4713} \\
			decrease or not    & 0.0837          & \textbf{0.4659} \\
			smallest change    & {0.0434} & \textbf{0.1413}          \\
			largest change     & {0.0724} & \textbf{0.1557}          \\ \midrule[0.6pt]
			Average Accuracy   & 0.0601          & \textbf{0.1969} \\
			Overall Accuracy   & 0.0559          & \textbf{0.1560} \\ \bottomrule[1pt]
	\end{tabular}}
\end{table}

To better understand what the model has learned for making decisions, we visualize attention maps of our model on some examples in Fig. \ref{att_vis}. It can be seen that the model learns to focus on the related changed regions to predict answers. From experimental results, we can conclude that more research efforts are needed to reach a satisfactory performance on the challenging CDVQA task. Specifically, more effective change analysis-based visual learning methods should be investigated. We also see that Transformer-based models have great potential for multi-temporal and multi-modal feature learning in CDVQA tasks. Additionally, self-supervised or unsupervised change detection methods need to be studied. How to obtain the semantic change information from multi-temporal data in an unsupervised manner is also an important research direction in CDVQA tasks.

\section{Conclusion}
\label{Conclusion}

To provide ordinary end users with flexible access to change information, we introduce a new task named CDVQA with natural language as output. This task takes multi-temporal aerial images and a natural question as inputs to predict the corresponding answer. To be specific, we create a new dataset, which contains 2,968 pairs of aerial images and more than 122,000 question-answer pairs. In addition, a baseline CDVQA model is devised, and different components of models are evaluated on the generated dataset. The experimental results outline possible problems that are needed to be addressed for the CDVQA task. This work also provides some useful insights for developing better CDVQA models, which are important for future research in this direction.

\section*{Acknowledgment}
\label{Acknowledgment}
We acknowledge the SECOND dataset contributors from Wuhan University and Ca' Foscari University of Venice.



\bibliographystyle{IEEEbib}

\bibliography{egbib}

\begin{thebibliography}{10}

\bibitem{saha2020building}
Sudipan Saha, Francesca Bovolo, and Lorenzo Bruzzone,
\newblock ``Building change detection in {VHR SAR} images via unsupervised deep
  transcoding,''
\newblock {\em IEEE Transactions on Geoscience and Remote Sensing}, vol. 59,
  no. 3, pp. 1917--1929, 2020.

\bibitem{ban2016change}
Yifang Ban and Osama Yousif,
\newblock ``Change detection techniques: A review,''
\newblock {\em Multitemporal Remote Sensing}, pp. 19--43, 2016.

\bibitem{you2020survey}
Yanan You, Jingyi Cao, and Wenli Zhou,
\newblock ``A survey of change detection methods based on remote sensing images
  for multi-source and multi-objective scenarios,''
\newblock {\em Remote Sensing}, vol. 12, no. 15, pp. 2460, 2020.

\bibitem{leenstra2021self}
Marrit Leenstra, Diego Marcos, Francesca Bovolo, and Devis Tuia,
\newblock ``Self-supervised pre-training enhances change detection in
  {S}entinel-2 imagery,''
\newblock {\em arXiv preprint arXiv:2101.08122}, 2021.

\bibitem{tian2014improving}
Jiaojiao Tian, Allan~Aasbjerg Nielsen, and Peter Reinartz,
\newblock ``Improving change detection in forest areas based on stereo
  panchromatic imagery using kernel {MNF},''
\newblock {\em IEEE Transactions on Geoscience and Remote Sensing}, vol. 52,
  no. 11, pp. 7130--7139, 2014.

\bibitem{baker2007change}
Corey Baker, Rick~L Lawrence, Clifford Montagne, and Duncan Patten,
\newblock ``Change detection of wetland ecosystems using landsat imagery and
  change vector analysis,''
\newblock {\em Wetlands}, vol. 27, no. 3, pp. 610--619, 2007.

\bibitem{schmitt2013wetland}
Andreas Schmitt and Brian Brisco,
\newblock ``Wetland monitoring using the curvelet-based change detection method
  on polarimetric {SAR} imagery,''
\newblock {\em Water}, vol. 5, no. 3, pp. 1036--1051, 2013.

\bibitem{washaya2018coherence}
Prosper Washaya, Timo Balz, and Bahaa Mohamadi,
\newblock ``Coherence change-detection with {S}entinel-1 for natural and
  anthropogenic disaster monitoring in urban areas,''
\newblock {\em Remote Sensing}, vol. 10, no. 7, pp. 1026, 2018.

\bibitem{qiao2020novel}
Huijiao Qiao, Xue Wan, Youchuan Wan, Shengyang Li, and Wanfeng Zhang,
\newblock ``A novel change detection method for natural disaster detection and
  segmentation from video sequence,''
\newblock {\em Sensors}, vol. 20, no. 18, pp. 5076, 2020.

\bibitem{olteanu2020use}
A-M Olteanu-Raimond, L~See, M~Schultz, G~Foody, M~Riffler, Tanja Gasber,
  Laurence Jolivet, Arnaud Le~Bris, Yann Meneroux, Lanfa Liu, et~al.,
\newblock ``Use of automated change detection and {VGI} sources for identifying
  and validating urban land use change,''
\newblock {\em Remote Sensing}, vol. 12, no. 7, pp. 1186, 2020.

\bibitem{mishra2014sensitivity}
Bhogendra Mishra and Junichi Susaki,
\newblock ``Sensitivity analysis for l-band polarimetric descriptors and fusion
  for urban land cover change detection,''
\newblock {\em IEEE Journal of selected topics in applied earth observations
  and remote sensing}, vol. 7, no. 10, pp. 4231--4242, 2014.

\bibitem{haack1998remote}
Barry Haack, James Wolf, and Richard English,
\newblock ``Remote sensing change detection of irrigated agriculture in
  {A}fghanistan,''
\newblock {\em Geocarto international}, vol. 13, no. 2, pp. 65--75, 1998.

\bibitem{liu2019review}
Sicong Liu, Daniele Marinelli, Lorenzo Bruzzone, and Francesca Bovolo,
\newblock ``A review of change detection in multitemporal hyperspectral images:
  Current techniques, applications, and challenges,''
\newblock {\em IEEE Geoscience and Remote Sensing Magazine}, vol. 7, no. 2, pp.
  140--158, 2019.

\bibitem{du2019unsupervised}
Bo~Du, Lixiang Ru, Chen Wu, and Liangpei Zhang,
\newblock ``Unsupervised deep slow feature analysis for change detection in
  multi-temporal remote sensing images,''
\newblock {\em IEEE Transactions on Geoscience and Remote Sensing}, vol. 57,
  no. 12, pp. 9976--9992, 2019.

\bibitem{li2019unsupervised}
Xuelong Li, Zhenghang Yuan, and Qi~Wang,
\newblock ``Unsupervised deep noise modeling for hyperspectral image change
  detection,''
\newblock {\em Remote Sensing}, vol. 11, no. 3, pp. 258, 2019.

\bibitem{lv2020object}
Zhiyong Lv, Tongfei Liu, and J{\'o}n~Atli Benediktsson,
\newblock ``Object-oriented key point vector distance for binary land cover
  change detection using {VHR} remote sensing images,''
\newblock {\em IEEE Transactions on Geoscience and Remote Sensing}, vol. 58,
  no. 9, pp. 6524--6533, 2020.

\bibitem{wang2018getnet}
Qi~Wang, Zhenghang Yuan, Qian Du, and Xuelong Li,
\newblock ``{GETNET}: A general end-to-end 2-{D} {CNN} framework for
  hyperspectral image change detection,''
\newblock {\em IEEE Transactions on Geoscience and Remote Sensing}, vol. 57,
  no. 1, pp. 3--13, 2018.

\bibitem{yang2020asymmetric}
Kunping Yang, Gui-Song Xia, Zicheng Liu, Bo~Du, Wen Yang, Marcello Pelillo, and
  Liangpei Zhang,
\newblock ``Asymmetric siamese networks for semantic change detection in aerial
  images,''
\newblock {\em IEEE Transactions on Geoscience and Remote Sensing}, vol. 60,
  2021.

\bibitem{ding2021bi}
Lei Ding, Haitao Guo, Sicong Liu, Lichao Mou, Jing Zhang, and Lorenzo Bruzzone,
\newblock ``Bi-temporal semantic reasoning for the semantic change detection in
  {HR} remote sensing images,''
\newblock {\em IEEE Transactions on Geoscience and Remote Sensing}, vol. 60,
  2022.

\bibitem{mogadala2019trends}
Aditya Mogadala, Marimuthu Kalimuthu, and Dietrich Klakow,
\newblock ``Trends in integration of vision and language research: A survey of
  tasks, datasets, and methods,''
\newblock {\em arXiv preprint arXiv:1907.09358}, 2019.

\bibitem{khamparia2020integrated}
Aditya Khamparia, Babita Pandey, Shrasti Tiwari, Deepak Gupta, Ashish Khanna,
  and Joel~JPC Rodrigues,
\newblock ``An integrated hybrid {CNN--RNN} model for visual description and
  generation of captions,''
\newblock {\em Circuits, Systems, and Signal Processing}, vol. 39, no. 2, pp.
  776--788, 2020.

\bibitem{ferraro2016visual}
Francis Ferraro, Nasrin Mostafazadeh, Ishan Misra, Aishwarya Agrawal, Jacob
  Devlin, Ross Girshick, Xiaodong He, Pushmeet Kohli, Dhruv Batra, C~Lawrence
  Zitnick, et~al.,
\newblock ``Visual storytelling,''
\newblock {\em arXiv preprint arXiv:1604.03968}, 2016.

\bibitem{antol2015vqa}
Stanislaw Antol, Aishwarya Agrawal, Jiasen Lu, Margaret Mitchell, Dhruv Batra,
  C~Lawrence Zitnick, and Devi Parikh,
\newblock ``{VQA}: Visual question answering,''
\newblock in {\em IEEE International Conference on Computer Vision (ICCV)},
  2015, pp. 2425--2433.

\bibitem{wu2017visual}
Qi~Wu, Damien Teney, Peng Wang, Chunhua Shen, Anthony Dick, and Anton van~den
  Hengel,
\newblock ``Visual question answering: A survey of methods and datasets,''
\newblock {\em Computer Vision and Image Understanding}, vol. 163, pp. 21--40,
  2017.

\bibitem{tu2021learning}
Tao Tu, Qing Ping, Govindarajan Thattai, Gokhan Tur, and Prem Natarajan,
\newblock ``Learning better visual dialog agents with pretrained
  visual-linguistic representation,''
\newblock in {\em IEEE/CVF Conference on Computer Vision and Pattern
  Recognition (CVPR)}, 2021, pp. 5622--5631.

\bibitem{zheng2021mutual}
Xiangtao Zheng, Binqiang Wang, Xingqian Du, and Xiaoqiang Lu,
\newblock ``Mutual attention inception network for remote sensing visual
  question answering,''
\newblock {\em IEEE Transactions on Geoscience and Remote Sensing}, 2021.

\bibitem{yuan2021self}
Zhenghang Yuan, Lichao Mou, and Xiao~Xiang Zhu,
\newblock ``Self-paced curriculum learning for visual question answering on
  remote sensing data,''
\newblock in {\em 2021 IEEE International Geoscience and Remote Sensing
  Symposium IGARSS}. IEEE, 2021, pp. 2999--3002.

\bibitem{qu2016deep}
Bo~Qu, Xuelong Li, Dacheng Tao, and Xiaoqiang Lu,
\newblock ``Deep semantic understanding of high resolution remote sensing
  image,''
\newblock in {\em 2016 International conference on computer, information and
  telecommunication systems (Cits)}. IEEE, 2016, pp. 1--5.

\bibitem{lu2017exploring}
Xiaoqiang Lu, Binqiang Wang, Xiangtao Zheng, and Xuelong Li,
\newblock ``Exploring models and data for remote sensing image caption
  generation,''
\newblock {\em IEEE Transactions on Geoscience and Remote Sensing}, vol. 56,
  no. 4, pp. 2183--2195, 2017.

\bibitem{liu2022remote}
Chenyang Liu, Rui Zhao, and Zhenwei Shi,
\newblock ``Remote-sensing image captioning based on multilayer aggregated
  transformer,''
\newblock {\em IEEE Geoscience and Remote Sensing Letters}, vol. 19, 2022.

\bibitem{lobry2020rs}
Sylvain Lobry, Diego Marcos, Jesse Murray, and Devis Tuia,
\newblock ``{RSVQA}: Visual question answering for remote sensing data,''
\newblock {\em IEEE Transactions on Geoscience and Remote Sensing}, vol. 58,
  no. 12, pp. 8555--8566, 2020.

\bibitem{yuan2022easy}
Zhenghang Yuan, Lichao Mou, Qi~Wang, and Xiao~Xiang Zhu,
\newblock ``From easy to hard: Learning language-guided curriculum for visual
  question answering on remote sensing data,''
\newblock {\em IEEE Transactions on Geoscience and Remote Sensing}, vol. 60,
  2022.

\bibitem{he2016deep}
Kaiming He, Xiangyu Zhang, Shaoqing Ren, and Jian Sun,
\newblock ``Deep residual learning for image recognition,''
\newblock in {\em IEEE/CVF Conference on Computer Vision and Pattern
  Recognition (CVPR)}, 2016, pp. 770--778.

\bibitem{vaswani2017attention}
Ashish Vaswani, Noam Shazeer, Niki Parmar, Jakob Uszkoreit, Llion Jones,
  Aidan~N Gomez, {\L}ukasz Kaiser, and Illia Polosukhin,
\newblock ``Attention is all you need,''
\newblock in {\em Advances in neural information processing systems (NeurIPS)},
  2017, pp. 5998--6008.

\bibitem{tetko2020state}
Igor~V Tetko, Pavel Karpov, Ruud Van~Deursen, and Guillaume Godin,
\newblock ``State-of-the-art augmented {NLP} transformer models for direct and
  single-step retrosynthesis,''
\newblock {\em Nature Communications}, vol. 11, no. 1, pp. 1--11, 2020.

\bibitem{liu2021swin}
Ze~Liu, Yutong Lin, Yue Cao, Han Hu, Yixuan Wei, Zheng Zhang, Stephen Lin, and
  Baining Guo,
\newblock ``Swin transformer: Hierarchical vision transformer using shifted
  windows,''
\newblock {\em arXiv preprint arXiv:2103.14030}, 2021.

\bibitem{dosovitskiy2020image}
Alexey Dosovitskiy, Lucas Beyer, Alexander Kolesnikov, Dirk Weissenborn,
  Xiaohua Zhai, Thomas Unterthiner, Mostafa Dehghani, Matthias Minderer, Georg
  Heigold, Sylvain Gelly, et~al.,
\newblock ``An image is worth 16x16 words: Transformers for image recognition
  at scale,''
\newblock in {\em International Conference on Learning Representations (ICLR)},
  2021.

\bibitem{carion2020end}
Nicolas Carion, Francisco Massa, Gabriel Synnaeve, Nicolas Usunier, Alexander
  Kirillov, and Sergey Zagoruyko,
\newblock ``End-to-end object detection with transformers,''
\newblock in {\em European Conference on Computer Vision (ECCV)}, 2020, pp.
  213--229.

\bibitem{xie2021segformer}
Enze Xie, Wenhai Wang, Zhiding Yu, Anima Anandkumar, Jose~M Alvarez, and Ping
  Luo,
\newblock ``Segformer: Simple and efficient design for semantic segmentation
  with transformers,''
\newblock {\em arXiv preprint arXiv:2105.15203}, 2021.

\bibitem{xiong2020msn}
Zhitong Xiong, Yuan Yuan, and Qi~Wang,
\newblock ``{MSN}: Modality separation networks for {RGB-D} scene
  recognition,''
\newblock {\em Neurocomputing}, vol. 373, pp. 81--89, 2020.

\bibitem{liu2021dual}
Yun Liu, Xiaoming Zhang, Qianyun Zhang, Chaozhuo Li, Feiran Huang, Xianghong
  Tang, and Zhoujun Li,
\newblock ``Dual self-attention with co-attention networks for visual question
  answering,''
\newblock {\em Pattern Recognition}, vol. 117, pp. 107956, 2021.

\bibitem{xiong2021ask}
Zhitong Xiong, Yuan Yuan, and Qi~Wang,
\newblock ``{ASK}: Adaptively selecting key local features for {RGB-D} scene
  recognition,''
\newblock {\em IEEE Transactions on Image Processing}, vol. 30, pp. 2722--2733,
  2021.

\bibitem{chappuis2021find}
Christel Chappuis, Sylvain Lobry, Benjamin Kellenberger, Bertrand~Le Saux, and
  Devis Tuia,
\newblock ``How to find a good image-text embedding for remote sensing visual
  question answering?,''
\newblock {\em arXiv preprint arXiv:2109.11848}, 2021.

\bibitem{kiros2015skip}
Ryan Kiros, Yukun Zhu, Russ~R Salakhutdinov, Richard Zemel, Raquel Urtasun,
  Antonio Torralba, and Sanja Fidler,
\newblock ``Skip-thought vectors,''
\newblock {\em Advances in neural information processing systems}, vol. 28,
  2015.

\bibitem{shao2021sunet}
Ruizhe Shao, Chun Du, Hao Chen, and Jun Li,
\newblock ``{SUNet}: Change detection for heterogeneous remote sensing images
  from satellite and {UAV} using a dual-channel fully convolution network,''
\newblock {\em Remote Sensing}, vol. 13, no. 18, pp. 3750, 2021.

\end{thebibliography}

\end{document}